\documentclass[pdflatex,sn-mathphys-num]{sn-jnl}

\usepackage{graphicx}%
\usepackage{multirow}%
\usepackage{amsmath,amssymb,amsfonts}%
\usepackage{amsthm}%
\usepackage{mathrsfs}%
\usepackage[title]{appendix}%
\usepackage{xcolor}%
\usepackage{textcomp}%
\usepackage{manyfoot}%
\usepackage{booktabs}%
\usepackage{algorithm}%
\usepackage{algorithmicx}%
\usepackage{algpseudocode}%
\usepackage{listings}%
\usepackage{makecell} 
\usepackage[table]{xcolor}%
\usepackage{xspace} %
\usepackage{rotating}%
\usepackage{amssymb}%
\usepackage{colortbl}
\usepackage{adjustbox} 
\usepackage{subcaption} 

\usepackage[draft]{todonotes}   
\usepackage[defaultcolor=red,commandnameprefix=always]{changes}

\theoremstyle{thmstyleone}%
\theoremstyle{thmstyletwo}%

\theoremstyle{thmstylethree}%

\begin{document}
\title[Article Title]{When to Trust the Answer: Question-Aligned Semantic Nearest Neighbor Entropy for Safer Surgical VQA}


\author*[1]{\fnm{Luca} \sur{Carlini}}\equalcont{These authors contributed equally to this work.}\email{luca.carlini@polimi.it}
\author*[1]{\fnm{Dennis Pierantozzi}}\equalcont{These authors contributed equally to this work.}\email{dennis.pierantozzi@mail.polimi.it}

\author[1]{\fnm{Mauro Orazio} \sur{Drago}}
\author[1]{\fnm{Chiara} \sur{Lena}}

\author[2]{\fnm{Cesare} \sur{Hassan}}

\author[1]{\fnm{Elena} \sur{De Momi}}

\author[3]{\fnm{Danail} \sur{Stoyanov}}

\author[3]{\fnm{Sophia} \sur{Bano}}

\author*[3,4]{\fnm{Mobarak I.} \sur{Hoque}} \email{mobarak.hoque@manchester.ac.uk}

\affil[1]{\orgdiv{Dipartimento di Elettronica, Informazione e Bioingegneria (DEIB)}, \orgname{Politecnico di Milano}, \country{Italy}}
\affil[2]{\orgdiv{IRCCS Humanitas Research Hospital},  
\country{Italy}}

\affil[3]{
\orgdiv{UCL Hawkes Institute and Department of Computer Science}, 
\orgname{University College London},
\orgaddress{\country{UK}}
}

\affil[4]{\orgdiv{Division of Informatics, Imaging and Data Science}, \orgname{University of Manchester}, \orgaddress{\country{UK}}}

\abstract{
\textbf{Purpose:} 
Safety and reliability are critical for deploying visual question answering (VQA) systems in surgery, where incorrect or ambiguous responses can cause patient harm. A key limitation of existing uncertainty estimation methods, such as Semantic Nearest Neighbor Entropy (SNNE), is that they do not explicitly account for the conditioning question. Consequently, these methods may assign high confidence to answers that are consistent but misaligned with the clinical question, particularly under variation in question phrasing.

\textbf{Methods:} 
We propose Question-Aligned Semantic Nearest Neighbor Entropy (QA-SNNE), a black-box uncertainty estimator that incorporates question–answer alignment into semantic entropy through bilateral gating. QA-SNNE measures uncertainty by weighting pairwise semantic similarities among sampled answers according to their relevance to the question, using embedding-based, entailment-based, or cross-encoder alignment strategies. To assess robustness under language variation, we construct an out-of-template rephrased version of a benchmark surgical VQA dataset, in which only the question wording is modified while images and ground-truth answers remain unchanged.

\textbf{Results:} 
We evaluate QA-SNNE on five VQA models across two benchmark surgical VQA datasets, considering both zero-shot and parameter-efficient fine-tuned (PEFT) settings, including out-of-template questions. QA-SNNE improves AUROC on EndoVis18-VQA for two of three zero-shot models in-template (e.g., +15\% for Llama3.2 and +21\% for Qwen2.5) and achieves up to +8\% AUROC improvement under out-of-template rephrasing, with mixed results on external validation.

\textbf{Conclusion:} 
By making uncertainty estimation explicitly question-aware, QA-SNNE provides a practical and model-agnostic safeguard for surgical VQA. Linking semantic uncertainty to question relevance enables more reliable identification of unsafe predictions and supports safer deployment under variation in question phrasing.

}

\keywords{Surgical VQA, Uncertainty estimation, Large Vision-Language Models, Semantic Entropy}



\maketitle

\section{Introduction}\label{sec1}

Minimally invasive and image-guided procedures demand rapid, reliable interpretation of complex visual scenes. Surgeons must reason over instrument motion, tissue appearance, and evolving anatomy while operating under time pressure and with limited field of view. Visual Question Answering (VQA) for surgery has emerged as a compelling paradigm for turning raw pixels into actionable, query-conditioned information that could support intraoperative decision-making and surgical training~\cite{seenivasan2022surgical}. 
In clinical deployment, accuracy is not enough: when uncertain, the system must default to safety, particularly when the model may misunderstand or fail to answer the question being asked.

Most existing surgical VQA studies optimize for utility~\cite{seenivasan2022surgical, he2024pitvqa, he2025pitvqa++} and only indirectly touch on safety. Two limitations recur. First, systems often lack explicit mechanisms to recognize and communicate uncertainty, to abstain, or to route queries to a human expert. Second, evaluations are commonly conducted under “in-template” conditions, where test questions closely mirror training phrasings, this setup encourages text-matching shortcuts and overestimates robustness to the linguistic drift that is routine in real clinical conversations. As a result, models may appear competent while remaining brittle to language variation, negation, or clinically subtle rewordings, and they may fail to calibrate confidence to the true likelihood of error.

Concurrently, progress in uncertainty estimation and Automatic Failure Detection (AFD) has introduced semantics-driven approaches for identifying unreliable outputs from Large Language Models (LLMs)~\cite{shorinwa2025survey} and Large Vision–Language Models (LVLMs)~\cite{liu2024survey}. Notably, Semantic Entropy (SE)~\cite{farquhar2024detecting} pioneered measuring uncertainty through semantic clustering of generated responses, moving beyond token-level probabilities to capture meaning-level consistency. More recently, Semantic Nearest Neighbor Entropy (SNNE)~\cite{nguyen2025beyond} refined this approach by computing pairwise semantic similarities without explicit clustering, offering computational advantages and improved discrimination. 

However, these methods remain question-agnostic and assess answer consistency without explicitly considering how well responses address the question that conditions the prediction, which can lead to overconfident uncertainty estimates when answers are mutually consistent but fail to satisfy the question.

In this paper, we address safer answer selection for surgical visual question answering (VQA) under language variation (out-of-template) drift and heterogeneous LVLM backbones by explicitly incorporating question alignment into the uncertainty estimation process. We introduce an out-of-template version of the EndoVis18-VQA~\cite{seenivasan2022surgical} dataset and design a Question-Aligned Semantic Nearest-Neighbor Entropy (QA-SNNE) hallucination detector that extends semantic entropy with question-aware scoring.

Our key contributions are as follows:

\begin{itemize}



\item[--] In this paper, we address safer answer selection for surgical visual question answering (VQA) under rephrase (out-of-template) drift and heterogeneous LVLM backbones by explicitly incorporating question alignment into the uncertainty estimation process, so that uncertainty reflects not only agreement among answers but also whether those answers satisfy the question being asked.

\item[--] We construct an out-of-template variant of EndoVis18-VQA, which will be publicly released with this paper, by rephrasing questions while strictly preserving clinical intent and the original answers. This resource complements in-template testing and offers a reproducible stress test for semantics-first generalization in surgical VQA, specifically under question reformulations.

\item[--] We conduct extensive experiments on five models covering both Parameter-Efficient Fine-Tuning (PEFT) and zero-shot LVLM backbones on the different templates plus an external validation. 
question-aligned QA-SNNE improves AUROC, especially under rephrase drift
Because of its black-box and output-only nature, our method generalizes cleanly across models and datasets, strengthening the safety and reliability of LVLM deployments in surgical settings.


\end{itemize}

\section{Methodology}
\subsection{EndoVis18-VQA: Out-of-template}

Language in the operating room is fluid: identical clinical intent is often expressed with different words, levels of explicitness, and local habits of speech. Template-constrained benchmarks can therefore overstate robustness, as models may learn to match familiar surface forms rather than ground their answers in the image. Our out-of-template evaluation targets this gap by testing the \emph{invariance} that should hold under semantically faithful rephrasing, an idea rooted in behavioral testing for NLP and complementary to distribution-shift stress tests in general VQA. Prior work shows that small lexical or structural edits can disrupt model predictions; surgical VQA systems should not be brittle in this way ~\cite{ribeiro2020beyond, agrawal2018don}.


Starting from the EndoVis18-VQA resource~\cite{seenivasan2022surgical}, we rephrased the 35 questions present in the original (in-template) (covering tool, location, action, and organ queries), and manually verified that each paraphrase preserved the original clinical intent, answer type, and referent. For the out-of-template variant, we keep every image, answer, and data split untouched and modify only the surface form of each question, yielding a drop-in replacement that isolates the effect of clinically realistic rephrasing drift without altering ground truth or imagery. The procedure is intentionally simple and transparent. We have rephrased each template to mirror how questions are naturally posed during procedures, frequently making intent explicit and resolving potential ambiguities in everyday shorthand, while preserving the answer type and referent. In keeping with the invariance principle, only the wording changed while images and answers remained identical. As illustrated in Table~\ref{tab:paraphrase_taxonomy}, each in-template question is paired with its out-of-template counterpart for the same frame; examples include reformulating “state of” to “function currently being performed” and clarifying “located” to “currently positioned within the surgical field,” while maintaining answer identity.



\begin{table}[t!]
\centering
\small
\setlength{\tabcolsep}{6pt}
\begin{tabular}{l r p{3.2cm} p{5.5cm}}
\toprule
\textbf{Type} & \textbf{\#Q} & \textbf{Original} & \textbf{Rephrased} \\
\midrule
Tool & 17 & What is the state of bipolar forceps? & What is the function currently being performed by the bipolar forceps during the surgical procedure? \\
Location & 16 & Where is clip applier located? & Where is the clip applier currently positioned within the surgical field? \\
Organ & 2 & What organ is being operated? & What specific abdominal organ is currently undergoing surgical intervention during the robotic-assisted procedure? \\
\bottomrule
\end{tabular}
\caption{Rephrasing taxonomy and counts over the $n{=}35$ questions (\#Q). Only wording changes; images, answers, and splits remain identical.}
\label{tab:paraphrase_taxonomy}
\end{table}




\subsection{Question-Aligned Semantic Nearest-Neighbor Entropy}


\textbf{Background:} 
Hallucination detection methods span three categories: uncertainty-based approaches infer errors from predictive uncertainty without extra supervision~\cite{li2024reference, farquhar2024detecting}; detector-based methods train classifiers on labeled hallucination data~\cite{zhang2024dhcp} and visual evidence-verification tests image-text faithfulness through input perturbations~\cite{yin2024woodpecker} such as VL-Uncertainty (VL-U)~\cite{zhang2024vl}. Uncertainty estimation is particularly attractive for its simplicity and black-box applicability~\cite{cossio2025comprehensive}. 
A recent milestone in this direction has been made by Semantic Entropy (SE) advancing beyond token-level metrics by measuring semantic neighborhood uncertainty~\cite{farquhar2024detecting}. 
Our approach builds upon Semantic Nearest Neighbor Entropy (SNNE)~\cite{nguyen2025beyond}, a new state of the art uncertainty estimation method which estimates uncertainty by computing pairwise similarities among sampled answers without requiring explicit clustering. Given a question $q$ and $n$ generated answers $\{a_1, \ldots, a_n\}$ sampled at high temperature, SNNE constructs a text similarity matrix $S^{\text{text}} \in \mathbb{R}^{n \times n}$ and computes uncertainty as:
\begin{equation}
    \text{SNNE}(q) = -\frac{1}{n} \sum_{i=1}^{n} \log \sum_{\substack{j=1 \\ j \neq i}}^{n} \exp\left(\frac{S^{\text{text}}_{ij}}{\tau}\right),
\end{equation}
where $\tau$ is a temperature parameter. Unlike discrete Semantic Entropy, SNNE naturally captures both intra-cluster similarity (when $a_i$ and $a_j$ are semantically equivalent) and inter-cluster dissimilarity through the continuous similarity function, avoiding the need for explicit clustering.

\noindent \textbf{Motivation:}
Extending SNNE to medical vision-language models reveals a critical tension: strong visual perturbations risk distorting diagnostic cues~\cite{ma2021understanding}, while weak perturbations are ignored by models that over-rely on language priors~\cite{favero2024multi, agrawal2018don}, exposing a gap in semantics-aware uncertainty methods that preserve clinical image fidelity. On the other hand standard uncertainty quantification methods for generative models often ignore the question context when assessing answer reliability. In medical visual question answering, however, the question provides strong structural priors over the answer space. For instance, ``Which tool...?'' implies a categorical choice from a finite set, ``Where...?'' indicates spatial localization, and ``What is the state of...?'' suggests classification over states or actions. We leverage this observation to develop a question-aligned uncertainty measure that explicitly incorporates question-answer alignment into the uncertainty estimation process.

\begin{figure*}[t!]
    \centering
    \includegraphics[width=0.95\textwidth]{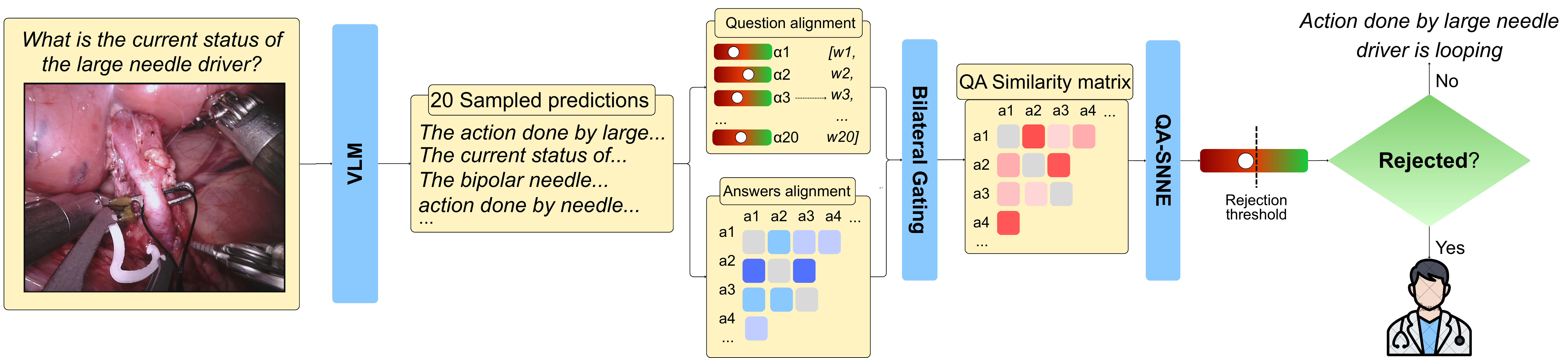}
    \caption{QA-SNNE Framework for Automatic Failure Detection.}
    \label{fig:qa_snne_framework}
\end{figure*}

\noindent \textbf{Question-Aligned Gating Mechanism:} We extend SNNE by incorporating question-answer alignment directly into the similarity matrix through a gating mechanism. The process consists of three steps:

\begin{enumerate}
    \item \textit{Compute alignment scores:} For each answer $a_i$, we compute an alignment score $\alpha_i \in \mathbb{R}$ that measures how well it addresses the question $q$ (see variants below).
    
    \item \textit{Convert to relevance weights:} The alignment scores are transformed into normalized relevance weights via softmax with sharpness parameter $\beta$:
    \begin{equation}
            w_i = \frac{\exp(\beta \cdot \alpha_i)}{\sum_{k=1}^{n} \exp(\beta \cdot \alpha_k)},
    \end{equation}
    where $\beta > 0$ controls the concentration of the distribution (default $\beta = 10$). Higher $\beta$ values produce sharper distinctions between well and poorly-aligned answers.
    
    \item \textit{Apply bilateral gating:} The similarity matrix is gated via row–column scaling:
    \begin{equation}
        S^{\text{QA}}_{ij} = w_i \cdot S^{\text{text}}_{ij} \cdot w_j = \text{diag}(\mathbf{w}) \cdot S^{\text{text}} \cdot \text{diag}(\mathbf{w}),
    \end{equation}
    where $\mathbf{w} = [w_1, \ldots, w_n]^\top$. This bilateral scaling ensures that pairwise similarities are down-weighted whenever \textit{either} answer has low alignment with the question.
\end{enumerate}

The gated similarity matrix $S^{\text{QA}}$ is used to compute QA-SNNE:
\begin{equation}
    \text{QA-SNNE}(q) = -\frac{1}{n} \sum_{i=1}^{n} \log \sum_{\substack{j=1 \\ j \neq i}}^{n} \exp\left(\frac{S^{\text{QA}}_{ij}}{\tau}\right).
\end{equation}

Answers with low alignment scores $\alpha_i$ and thus low weights $w_i$ contribute less to the similarities that drive the entropy estimate. This keeps QA-SNNE sensitive to semantic consistency while enforcing question relevance, balancing the benefits of SNNE with protection against coherent but off-question answer clusters.

With high-temperature sampling, models often produce a mix of answers that address the question and answers that are generic, evasive, or effectively respond to a different interpretation. If these off-topic samples are treated as equal modes, uncertainty can rise due to variability that is unrelated to the clinical query, which weakens separation between correct and incorrect answers. The gating step reduces the influence of misaligned samples so QA-SNNE measures disagreement among question-relevant answers. Misalignment is still a safety signal: when the model frequently goes off-topic, fewer aligned samples remain and they are typically less consistent, which increases QA-SNNE, and the alignment scores themselves indicate that the model did not answer the question.

\noindent \textbf{Question-Answer Alignment Variants:} We present and explore three methods ("Embedding", "Entailment", "Cross-Encoder") for computing the alignment scores $\alpha_i$.
All three variants produce unbounded alignment scores $\alpha_i \in \mathbb{R}$, which are then normalized via the softmax transformation (Step 2) before gating the similarity matrix.

\begin{description}
    \item[Embedding-based (Emb):] We encode each question and answer using domain-adapted sentence embeddings and compute alignment as the cosine similarity between their representations: $\alpha_i = \text{cos}(\mathbf{e}_q, \mathbf{e}_{a_i})$.
    
    \item[Entailment-based (Ent):] We employ a natural language inference model to assess bidirectional semantic compatibility. For each answer $a_i$, we compute entailment and contradiction logits in both directions ($q \to a_i$ and $a_i \to q$), then combine them as:
    \begin{equation}
            \alpha_i = \gamma \left(\ell_{\text{ent}}^{q\to a_i} + \ell_{\text{ent}}^{a_i\to q}\right) - \lambda\left(\ell_{\text{con}}^{q\to a_i} + \ell_{\text{con}}^{a_i\to q}\right),
    \end{equation}
    where $\gamma$ weights entailment evidence and $\lambda$ penalizes contradictions.
    This formulation rewards mutual entailment while penalizing contradictions, capturing logical consistency between question and answer.
    
    \item[Cross-encoder-based (CrossE):] We apply a cross-encoder re-ranker that jointly processes question-answer pairs $(q, a_i)$. Unlike bi-encoder approaches, this model performs full cross-attention over both sequences, yielding relevance logits $\alpha_i$ that directly measure answer appropriateness without relying on independent encodings.
\end{description}

\noindent\textbf{Hallucination Detection.} While our QA-SNNE are continuous, we evaluate them against binary hallucination labels using threshold-based classification. Given uncertainty scores $\{u_1, \ldots, u_n\}$, we set a threshold $\theta^*$ and classify answers as hallucinatory if $u_i \geq \theta^*$, otherwise as grounded. This enables direct comparison with existing hallucination detection benchmarks 
preserving continuous uncertainty signal for additional analyses such as selective prediction and calibration curves.

\section{Experiments and Results}\label{sec:experiments}
\subsection{Datasets}
\textbf{EndoVis18-VQA (in-template).} We use the standard EndoVis18-VQA~\cite{seenivasan2022surgical} dataset derived from MICCAI EndoVis 2018 nephrectomy videos with question templates covering tool, location, action and organ queries. We have considered only the validation sequences that comprise 2,754 image–question pairs. 
\textbf{EndoVis18-VQA (out-of-template, ours).} The out-of-template split mirrors the in-template size (2,754 pairs) and contains the questions rephrased as discussed in section 2.1. 
\textbf{Open-ended PitVQA (external).} We use the open-ended pituitary surgery VQA dataset~\cite{he2025pitvqa++} for external validation, consisting of procedural images and 4766 diverse QA pairs.

\subsection{Implementation details} 
Our QA-SNNE method is implemented with PyTorch. The uncertainty estimator operates as a black-box post-hoc module over model outputs. We follow a three-stage protocol: (i) generate a single answer at low temperature ($T = 0.1$), encouraging the model to produce a high-confidence response, and compare it with the ground truth dataset annotation using ROUGE-L, with a fixed threshold of 0.5 to derive ground-truth hallucination labels. This threshold was selected to balance coverage and performance, avoiding overly strict matching while still penalizing clearly incorrect generations; (ii) draw $n = 20$ diverse samples at high temperature ($T = 1.0$, top-$k = 50$, top-$p = 0.9$) to compute uncertainty from the sampled distribution; and (iii) apply a threshold of $-3.5$ to detect hallucinations and score accuracy against labels from step (i). The $-3.5$ threshold was chosen based on evaluation on a subset of the training data and then kept fixed across all experiments.

We evaluate three variants: (a) embedding-based (PubMed-adapted sentence embeddings; cosine~\cite{deka2022improved}), (b) bidirectional NLI (DeBERTa-large-MNLI~\cite{he2020deberta}; entailment/contradiction weighting), and (c) cross-encoder re-ranking (BGE-reranker-large~\cite{chen2024bge}). All models are obtained from official Hugging Face repositories. We use $\beta = 10$ for softmax sharpness and ROUGE-L for base similarity before bilateral gating.

We use as state-of-the-art baselines the black-box variants of Discrete Semantic Entropy (DSE)~\cite{farquhar2024detecting}, SNNE~\cite{nguyen2025beyond}, and VL-U~\cite{zhang2024vl}, implemented following their official repositories. In line with our deployment setting, we restrict comparisons to post-hoc estimators that operate only on generated answers and thus do not require access to logits, internal activations, or additional training data. For hallucination detection, we apply the same threshold for all semantic entropy–based methods, while using the method-specific threshold for VL-U as recommended in the original work~\cite{zhang2024vl}. For fairness, all comparative baselines, including SurgicalGPT and PitVQA, are retrained using their official repositories on the EndoVis18-VQA in-template dataset. For LVLM backbones we use Llama-3.2-11B-Vision-Instruct~\cite{dubey2024llama}, MedGemma-4B-it~\cite{sellergren2025medgemma}, and Qwen2.5-VL-3B-Instruct~\cite{Qwen-VL} at inference with zero-shot modality injection via a prompt describing the surgical environment. Experiments are conducted on NVIDIA A6000 and L40S GPUs.

\subsection{Results}

\subsubsection{Quantitative results}

\begin{table*}[ht!]
\centering
\small
\caption{\textbf{Utility and Safety Metrics Across Validation Sets.} We report BLEU, ROUGE-L, METEOR and AUROC. Higher is better for all the metrics. Bold indicates the best within each column-block for utility and best method within the row-block for safety, underlined the second best for safety metric.}
\resizebox{\textwidth}{!}{%
\begin{tabular}{llccc|cccccc}
\toprule
& & \multicolumn{3}{c}{\textbf{Utility}} & \multicolumn{6}{c}{\textbf{Safety (AUROC)}} \\
\cmidrule(lr){3-5} \cmidrule(lr){6-11}
& \textbf{Model} & \textbf{BLEU} & \textbf{ROU-L} & \textbf{MET} & \textbf{DSE}~\cite{farquhar2024detecting} & \textbf{SNNE}~\cite{nguyen2025beyond} & \textbf{VL-U}~\cite{zhang2024vl} & \multicolumn{3}{c}{\textbf{QA-SNNE (Ours)}} \\
\cmidrule(lr){9-11}
& & & & & & & & \textit{Emb} & \textit{Ent} & \textit{CrossE} \\
\midrule
\multicolumn{11}{c}{\textbf{(a) EndoVis18-VQA validation (In-template)}} \\
\midrule
\multirow{3}{*}{\rotatebox{90}{Zero-shot}}
 & Llama3.2~\cite{dubey2024llama} & 0.239 & 0.444 & 0.503 & 0.572 & 0.510 & \underline{0.685} & 0.527 & 0.551 & \textbf{0.789} \\
 & MedGemma3.0~\cite{sellergren2025medgemma} & 0.079 & 0.232 & 0.279 & 0.544 & \textbf{0.721} & 0.501 & \underline{0.690} & 0.618 & 0.530 \\
 & Qwen2.5~\cite{Qwen-VL} & 0.269 & 0.387 & 0.413 & 0.532 & 0.536 & \underline{0.656} & 0.505 & 0.559 & \textbf{0.794} \\[1mm]
\midrule
\multirow{2}{*}{\rotatebox{90}{Peft}}
 & PitVQA~\cite{he2024pitvqa} & \textbf{0.836} & \textbf{0.784} & \textbf{0.799} & 0.766 & \underline{0.886} & 0.500 & \textbf{0.914} & 0.879 & 0.849 \\
 & SurgicalGPT~\cite{seenivasan2023surgicalgpt} & 0.620 & 0.585 & 0.579 & \underline{0.958} & 0.893 & 0.500 & \textbf{0.993} & 0.507 & 0.632 \\
\midrule
\multicolumn{11}{c}{\textbf{(b) EndoVis18-VQA validation (Out-of-template)}} \\
\midrule
\multirow{3}{*}[-7pt]{\rotatebox{90}{Zero-shot}}
 & Llama3.2~\cite{dubey2024llama} & 0.201 & 0.337 & 0.357 & \textbf{0.673} & 0.638 & 0.532 & \underline{0.663} & 0.527 & 0.528 \\
 & MedGemma3.0~\cite{sellergren2025medgemma} & 0.167 & 0.267 & 0.272 & 0.507 & \underline{0.798} & 0.561 & \textbf{0.816} & 0.511 & 0.699 \\
 & Qwen2.5~\cite{Qwen-VL} & 0.280 & 0.325 & 0.337 & 0.553 & 0.554 & 0.556 & \textbf{0.601} & \underline{0.598} & 0.540 \\
 & PitVQA~\cite{he2024pitvqa} & \textbf{0.474} & \textbf{0.468} & \textbf{0.454} & 0.547 & 0.588 & 0.500 & \textbf{0.760} & 0.553 & \underline{0.739} \\
 & SurgicalGPT~\cite{seenivasan2023surgicalgpt} & 0.373 & 0.439 & 0.449 & \underline{0.617} & \textbf{0.795} & 0.500 & 0.502 & 0.546 & 0.505 \\
\midrule
\multicolumn{11}{c}{\textbf{(c) Open-ended PitVQA (External Validation)}} \\
\midrule
\multirow{3}{*}[-7pt]{\rotatebox{90}{Zero-shot}}
 & Llama3.2~\cite{dubey2024llama} & 0.124 & 0.210 & 0.300 & 0.819 & \textbf{0.937} & 0.540 & \underline{0.834} & 0.555 & 0.527 \\
 & MedGemma3.0~\cite{sellergren2025medgemma} & 0.263 & 0.321 & 0.359 & 0.560 & 0.540 & \underline{0.687} & \textbf{0.755} & 0.538 & 0.636 \\
 & Qwen2.5~\cite{Qwen-VL} & \textbf{0.441} & \textbf{0.588} & \textbf{0.632} & 0.540 & \underline{0.682} & \textbf{0.721} & 0.587 & 0.515 & 0.617 \\
 & PitVQA~\cite{he2024pitvqa} & 0.135 & 0.114 & 0.050 & 0.888 & \textbf{0.946} & 0.691 & \underline{0.926} & 0.587 & 0.504 \\
 & SurgicalGPT~\cite{seenivasan2023surgicalgpt} & 0.415 & 0.378 & 0.301 & \textbf{0.904} & 0.746 & 0.569 & \underline{0.881} & 0.790 & 0.830 \\
\bottomrule
\end{tabular}%
}
\label{tab:combined_metrics}
\end{table*}

\begin{figure}[t!]
\centering
\includegraphics[width=0.49\linewidth]{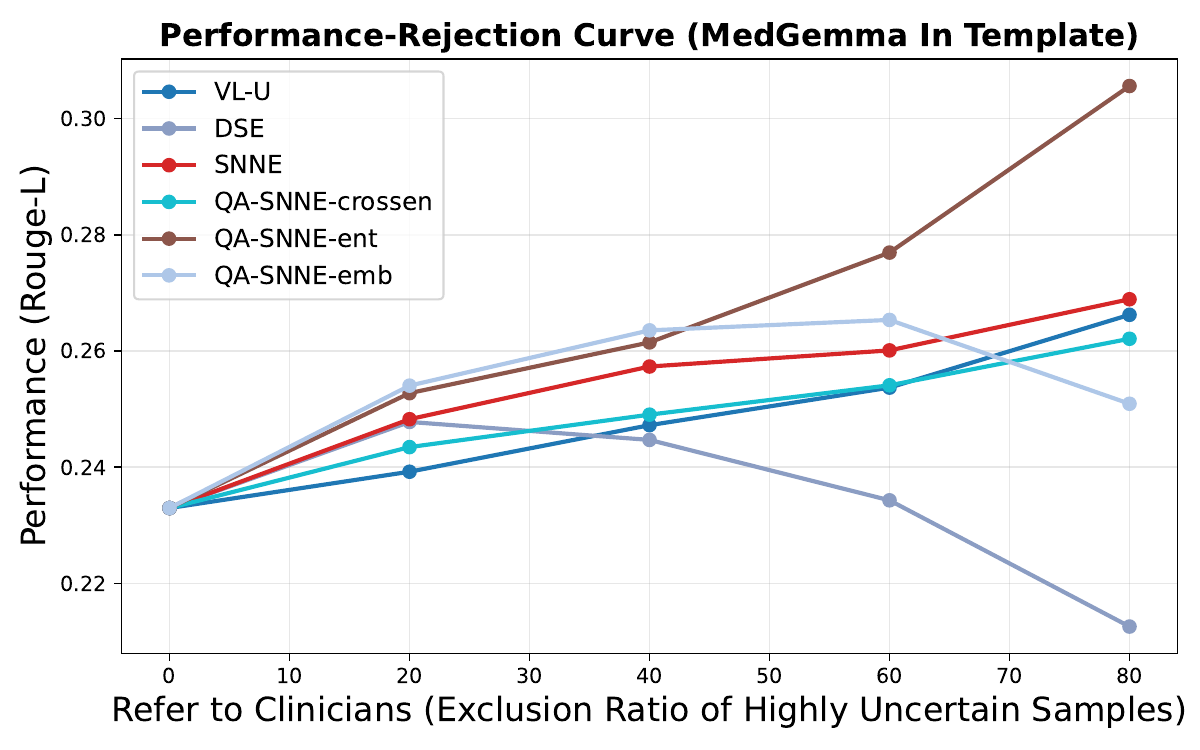}
\hfill
\includegraphics[width=0.49\linewidth]{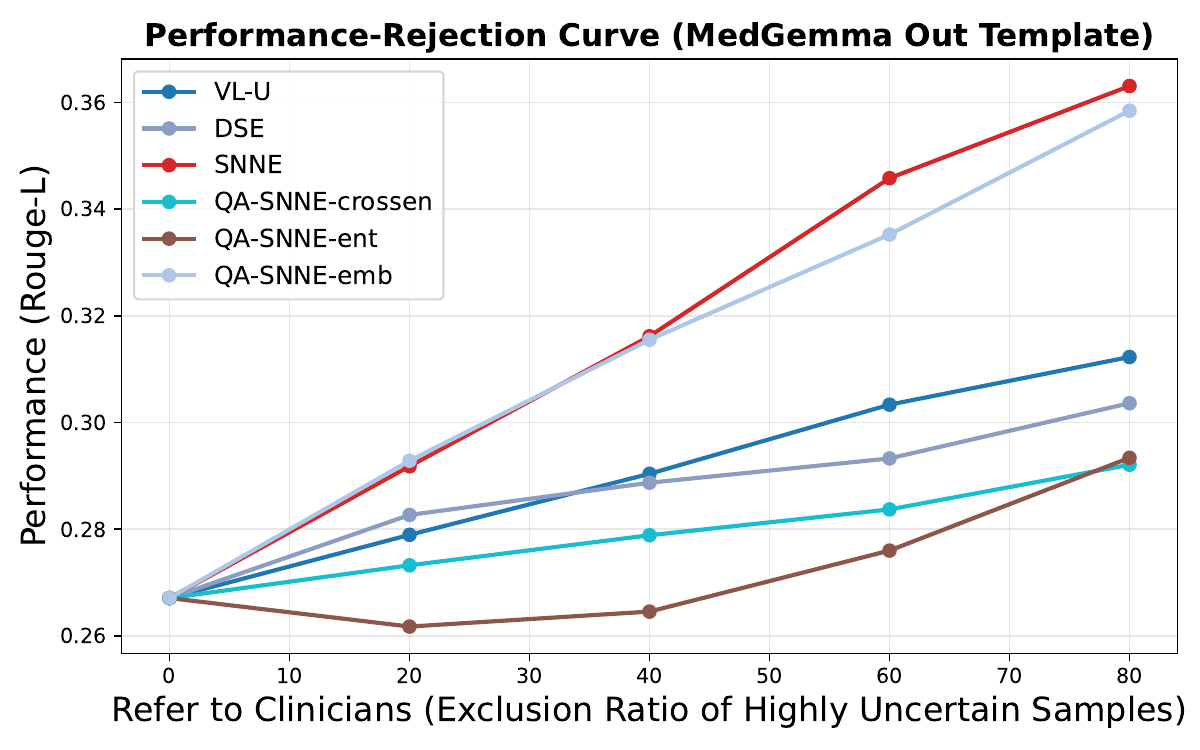}
\caption{PRC for MedGemma in template (left) and Out-of-template (right)}
\label{fig:pr_curves}
\end{figure}

Table~\ref{tab:combined_metrics} shows a fundamental trade-off between specialization and robustness. On in-template data, 
PEFT models achieve superior utility with PitVQA leads with BLEU/ROUGE-L/METEOR scores substantially outperforming zero-shot systems, demonstrating the value of domain adaptation when queries match training patterns. Under linguistic drift, PEFT models degrade: SurgicalGPT's BLEU drops from 0.620 to 0.373 on out-of-template rephrased, while Qwen2.5 maintains stability (from 0.269 to 0.280). On external PitVQA data, this reverses, Qwen2.5 leads (0.441 BLEU) while fine-tuned PitVQA drops (0.135). PEFT optimizes for narrow distributions but shows fragility under rephrase or domain shift; zero-shot models reduce peak accuracy for broader generalization.
Using Area Under the ROC Curve (AUROC) as our primary metric, QA-SNNE improves in most settings. On in-template data, zero-shot models improve substantially: Llama3.2 advances from 0.685 (VL-U) to 0.789 (+15\%), Qwen2.5 from 0.656 to 0.794 (+21\%). PEFT models also benefit: PitVQA reaches 0.914 (versus 0.886 SNNE), SurgicalGPT 0.993 (versus 0.893). Under out-of-template stress, gains persist: MedGemma improves from 0.798 to 0.816, Qwen2.5 from 0.554 to 0.601. External validation shows MedGemma achieving 0.755 where alternatives for the same model struggle below 0.700. 
Performance Rejection Curves (PRC) (Fig.~\ref{fig:pr_curves}) confirm QA-SNNE variants maintain superior ROUGE-L scores while selectively abstaining on high-uncertainty predictions, essential for safe clinical deployment.

\begin{table*}[t!]
\centering
\caption{\textbf{Accuracy across validation sets.}
Binary safety hallucination detection. Bold numbers denote the best method within each block, underlined is the second best.}
\setlength{\tabcolsep}{13pt}
\resizebox{\textwidth}{!}{%
\begin{tabular}{llccccc}
\toprule
& \textbf{Model} & \textbf{SNNE}~\cite{nguyen2025beyond} & \textbf{VL-U}~\cite{zhang2024vl} & \multicolumn{3}{c}{\textbf{QA-SNNE (Ours)}} \\
\cmidrule(lr){5-7}
 & & & & \textit{Emb} & \textit{Ent} & \textit{CrossE} \\
\midrule
\multicolumn{7}{c}{\textbf{(a) EndoVis18-VQA validation (In-template)}} \\
\midrule
\multirow{3}{*}{\rotatebox{90}{Zero-shot}}
 & Llama3.2~~\cite{dubey2024llama}       & 0.56 & \textbf{0.74} & 0.67 & 0.69 & \underline{0.70} \\
 & MedGemma3.0~\cite{sellergren2025medgemma}    & 0.22 & 0.81 & \underline{0.87} & \textbf{0.98} & \textbf{0.98} \\
 & Qwen2.5~\cite{Qwen-VL}        & 0.20 & \textbf{0.79} & 0.20 & \underline{0.67} & 0.54 \\ [1mm]
\midrule
\multirow{2}{*}{\rotatebox{90}{Peft}}
 & PitVQA~\cite{he2024pitvqa}      & \textbf{0.98} & \textbf{0.98} & \underline{0.85} & 0.01 & 0.51 \\
 & SurgicalGPT~\cite{seenivasan2023surgicalgpt} & \underline{0.82} & 0.39 & \textbf{0.84} & 0.35 & 0.35 \\
\midrule
\multicolumn{7}{c}{\textbf{(b) EndoVis18-VQA validation (Out-of-template)}} \\
\midrule
\multirow{3}{*}[-7pt]{\rotatebox{90}{Zero-shot}}
 & Llama3.2~\cite{dubey2024llama}       & 0.74 & 0.85 & 0.84 & \underline{0.96} & \textbf{0.97} \\
 & MedGemma3.0~\cite{sellergren2025medgemma}    & 0.17 & 0.76 & 0.77 & \textbf{0.98} & \underline{0.97} \\
 & Qwen2.5~\cite{Qwen-VL}        & 0.31 & 0.83 & \underline{0.87} & \textbf{0.93} & \textbf{0.93} \\
 & PitVQA~\cite{he2024pitvqa}     & 0.35 & 0.35 & \textbf{0.68} & 0.64 & \underline{0.67} \\
 & SurgicalGPT~\cite{seenivasan2023surgicalgpt} & 0.73 & 0.64 & \textbf{0.87} & 0.84 & \underline{0.85} \\
\midrule
\multicolumn{7}{c}{\textbf{(c) Open-ended PitVQA (External Validation)}} \\
\midrule
\multirow{3}{*}[-7pt]{\rotatebox{90}{Zero-shot}}
 & Llama3.2~\cite{dubey2024llama}       & \underline{0.91} & 0.73 & \textbf{0.92} & 0.74 & 0.79 \\
 & MedGemma3.0~\cite{sellergren2025medgemma}    & 0.48 & 0.73 & \textbf{0.77} & \underline{0.74} & \underline{0.74} \\
 & Qwen2.5~\cite{Qwen-VL}        & \underline{0.44} & \textbf{0.54} & 0.31 & 0.30 & 0.29 \\
 & PitVQA~\cite{he2024pitvqa}      & 0.27 & \underline{0.93} & 0.28 & \textbf{0.96} & \textbf{0.96} \\
 & SurgicalGPT~\cite{seenivasan2023surgicalgpt} & 0.17 & 0.64 & \underline{0.66} & \textbf{0.83} & \textbf{0.83} \\
\bottomrule
\end{tabular}%
}
\label{tab:accuracy_metrics}
\end{table*}

When converting continuous uncertainty scores to binary hallucination classifications using fixed thresholds (SNNE = -3.5, QA-SNNE = -3.5, VL-U = 1.0), entailment-based QA-SNNE demonstrates exceptional performance under linguistic drift (Table~\ref{tab:accuracy_metrics}). On the out-of-template split, where rephrased questions stress semantic robustness, our method achieves near-perfect accuracy for zero-shot models: 0.96 for Llama3.2 (versus 0.74 SNNE, 0.85 VL-U), 0.98 for MedGemma (versus 0.17 SNNE, 0.76 VL-U), and 0.93 for Qwen2.5 (versus 0.31 SNNE, 0.83 VL-U).
PEFT models reveal more complex behavior. On in-template data, certain QA-SNNE variants exhibit brittleness. PEFT models show a more model-dependent pattern. In particular, the discrete, thresholded accuracy can vary sharply across QA-SNNE variants, even when AUROC remains competitive. For PitVQA, the entailment-based variant collapses to 0.01 accuracy on the in-template split, whereas the same variant attains 0.35 on SurgicalGPT. A similar instability is observed for the cross-encoder variant. This behavior suggests that, for heavily fine-tuned models evaluated on highly templated questions, the high-temperature samples can become overly concentrated and linguistically homogeneous. In this regime, the alignment model may assign nearly identical (or saturated) relevance scores across samples, producing a gated similarity structure that interacts poorly with a fixed operating threshold. In other words, the failure here is not that entailment is uninformative in principle, but that the combination of limited answer diversity, alignment-score saturation, and a single global threshold can lead to miscalibrated binary decisions for some PEFT settings.
However, under out-of-template stress, question-aware uncertainty partially recovers: PitVQA achieves 0.64-0.68 across QA-SNNE variants (versus 0.35 for SNNE/VL-U), while SurgicalGPT reaches 0.84-0.87 (versus 0.73 SNNE, 0.64 VL-U). Entailment is strongest for zero-shot models under out-of-template drift, while Embedding often performs best for PEFT models, CrossE can be best for some zero-shot backbones.

\subsection{Qualitative results}
\label{subsec:qualitative}

On the EndoVis18 VQA out of template split, Table~\ref{tab:qual_snne_qasnne} shows that MedGemma3.0, Qwen2.5, and SurgicalGPT produce correct predictions with low uncertainty, while Llama3.2 exhibits a conservative false alarm where both methods abstain despite a correct answer, consistent with high sample diversity. We also observe a divergent failure mode on PitVQA, where a location query is answered with an action description and only QA-SNNE Ent triggers abstention. This indicates that question aligned uncertainty can surface off question generations that still appear self consistent to SNNE and would otherwise be treated as reliable.

\begin{table*}[t!]
\centering
\caption{\textbf{Qualitative behavior on EndoVis18 VQA out of template questions.}
One example per model.
We flag as hallucination when the uncertainty score satisfies $u \ge \theta$ with $\theta = -3.5$.
Bold predictions match the ground truth.
Bold uncertainty values indicate that the method decision matches correctness, meaning pass for correct predictions and flag for incorrect predictions.}
\label{tab:qual_snne_qasnne}
\scriptsize
\setlength{\tabcolsep}{4pt}
\renewcommand{\arraystretch}{1.45}
\resizebox{\textwidth}{!}{%
\begin{tabular}{p{2.0cm} p{5.2cm} p{5.0cm} p{5.6cm} c c}
\toprule
\textbf{Model} & \textbf{Question} & \textbf{Ground truth} & \textbf{Prediction} & \textbf{SNNE} & \textbf{QA-SNNE Ent} \\
\midrule
Llama3.2 &
Where is the large needle driver currently located? &
large needle driver is located at right top &
\textbf{The large needle driver is located in the upper right corner of the image.} &
-3.41 & -3.23 \\[1.0ex]

MedGemma3.0 &
What is the current status of the large needle driver? &
action done by large needle driver is looping &
\textbf{Yes, the large needle driver is being used.} &
\textbf{-3.89} & \textbf{-3.81} \\[1.0ex]

Qwen2.5 &
What is the current status of the large needle driver? &
action done by large needle driver is idle &
\textbf{The large needle driver is not in use.} &
\textbf{-3.75} & \textbf{-3.58} \\[1.0ex]

PitVQA &
Where are the prograsp forceps currently located? &
prograsp forceps is located at left bottom &
action done by prograsp forceps is contraction. &
-3.56 &
\textbf{-3.14} \\[1.0ex]

SurgicalGPT &
Where is the bipolar forceps currently located? &
bipolar forceps is located at left top &
\textbf{bipolar forceps is located at left top} &
\textbf{-3.81} & \textbf{-3.73} \\
\bottomrule
\end{tabular}%
}
\end{table*}

\section{Discussion}

Our main finding is that uncertainty should be question-aware to support safer surgical VQA. Question-agnostic semantic entropy methods can appear confident when sampled answers are mutually consistent but do not answer the question. This failure becomes more frequent under rephrase drift. QA-SNNE addresses it by weighting semantic agreement with question--answer alignment, so the score reflects both consistency and relevance.

The effect of QA-SNNE is regime-dependent. In in-template settings, many models already generate mostly on-question samples, so SNNE already captures the relevant semantic structure and QA-SNNE often yields limited additional gain. In out-of-template settings, sampled answers more often include fluent but only loosely related responses. Here SNNE can be over-confident because it rewards agreement without checking that the agreement is about the question. QA-SNNE corrects this by discounting off-question clusters, which explains why improvements are strongest under question drift and smaller when drift is absent.

The three variants differ in how they estimate alignment. The embedding-based variant is efficient and often strong for ranking (AUROC), but can be less discriminative when an answer is topically similar yet wrong. The cross-encoder variant can improve AUROC when fine-grained matching matters, but is less reliable when the dominant error is a shift in question interpretation. The entailment-based variant is most suitable when uncertainty must trigger a discrete action with a fixed threshold, because it tests logical compatibility between the question and the answer (support vs. conflict), which is particularly helpful under rephrasing.

A further reason for the observed differences across variants is that the underlying LVLMs generate qualitatively different answer sets under high-temperature sampling. Some models produce short categorical answers with limited lexical variation, while others generate longer, more conversational outputs that include hedging or extra context. These differences change the geometry of the answer similarity matrix and the behavior of the alignment estimator. For example, embedding similarity is sensitive to surface lexical overlap and may under-separate short answers that share tokens, while entailment can be sensitive to verbosity and implicit assumptions, and cross-encoders can over-reward stylistic plausibility when outputs are long and fluent. As a consequence, the same alignment strategy can behave differently depending on whether the model produces compact label-like answers or free-form explanations, and depending on how often sampling introduces off-question but internally consistent responses.

Finally, variant stability can degrade for strongly fine-tuned PEFT models on highly templated questions. These models may produce low-diversity high-temperature samples (near-duplicates), leading to nearly uniform alignment weights and making the gated similarity matrix close to SNNE. In this regime, small alignment-model biases can affect thresholded decisions even when ranking remains reasonable. 

For deployment, we therefore recommend selecting the variant based on the target use: embedding or cross-encoder for ranking-based review, and entailment for thresholded abstention under realistic language variation.

A limitation of our work is that our out-of-template EndoVis18-VQA split applies a single validated rephrasing per question template, which provides a controlled stress test that isolates the effect of wording changes while keeping images, answers, and splits identical. However, this one-to-one paraphrasing does not capture the full linguistic variability observed in real operating-room interactions, such as multiple alternative phrasings, ellipsis, negation, disfluencies, or varying levels of clinical specificity. As a result, the reported robustness gains should be interpreted as evidence under a constrained form of linguistic drift, and future work should extend the benchmark with richer multi-paraphrase and more diverse rewording strategies.

\section{Conclusion}

We studied safer surgical VQA under clinically realistic rephrase drift and showed that uncertainty estimation should be conditioned on the question, not only on answer agreement. Without question awareness, semantic entropy methods can become overconfident when answers are mutually consistent but misaligned with the intended query. QA-SNNE addresses this by incorporating question--answer alignment into SNNE, producing uncertainty scores that better track whether the model is answering the question being asked.

Across alignment strategies, we find complementary strengths, and no single variant is uniformly optimal. Embedding-based alignment is a competitive and efficient option for ranking-based safety evaluation, while cross-encoder alignment can yield strong AUROC when fine-grained question--answer matching is required. For fixed-threshold hallucination detection under rephrase drift, entailment-based QA-SNNE is the most dependable because it consistently achieves the highest or tied-highest accuracy for zero-shot models on out-of-template questions, indicating clearer separation between safe and unsafe predictions at a practical operating point. At the same time, the added alignment stage can sometimes be neutral or even harmful when answers are already well aligned or when the alignment estimator is mismatched to the model or dataset, which helps explain why QA-SNNE does not always outperform the base SNNE. Taken together, these results support using question-aligned uncertainty as a practical, model-agnostic safeguard for surgical VQA deployed under natural variation in how clinicians phrase questions.

\section*{Statements and Declarations}
\bmhead{Author Contributions} 
Luca Carlini and Dennis Pierantozzi contributed equally to this work and share first authorship.
\bmhead{Competing Interests} 
Luca Carlini, Dennis Pierantozzi, Mauro Orazio Drago, Chiara Lena, Cesare Hassan, Elena De Momi, Danail Stoyanov, Sophia Bano and Mobarak I. Hoque declare that they have no competing financial or non-financial interests related to this work.

\bmhead{Ethics Approval} 
This article does not contain any new studies with human participants or animals performed by any of the authors. All analyses were conducted on previously published, fully anonymized surgical video datasets (EndoVis18-VQA and PitVQA), for which ethical approval was obtained by the original data providers.

\bmhead{Informed Consent}
No new data involving human participants were collected for this study. Informed consent was obtained by the original investigators at the time of data acquisition, where applicable; therefore, additional informed consent for this secondary analysis was not required.

\bmhead{Fundings}

This work was supported by the Multilayered Urban Sustainability Action (MUSA) project (ECS00000037), funded by the European Union – NextGenerationEU under the National Recovery and Resilience Plan (NRRP); the ANTHEM project, funded by the National Plan for NRRP Complementary Investments (CUP: B53C22006700001); the Engineering and Physical Sciences Research Council (EPSRC) [EP/W00805X/1; UKRI145; EP/Y01958X/1]; the Wellcome/EPSRC Centre for Interventional and Surgical Sciences (WEISS) [203145/Z/16/Z]; and the Department for Science, Innovation and Technology (DSIT) and the Royal Academy of Engineering under the Chair in Emerging Technologies programme. For the purpose of open access, the author has applied a CC BY public copyright license to any Author Accepted Manuscript arising from this submission.

\bmhead{Code availability}
The source code of this work, along with the Endovis18-VQA out-of-template dataset, is available at \href{https://github.com/DennisPierantozzi/QA-SNNE}{GitHub repository}.

\bibliography{sn-bibliography}

\end{document}